\documentclass[letterpaper,10pt,conference]{ieeeconf}
\IEEEoverridecommandlockouts
\usepackage{cite}
\usepackage{amsmath,amssymb,amsfonts}
\usepackage{algorithmic}
\usepackage{graphicx}
\usepackage{textcomp}
\usepackage{xcolor}
\usepackage{booktabs}
\usepackage{multirow}
\usepackage{float}
\usepackage{listings}
\usepackage[most]{tcolorbox}
\lstdefinestyle{appendixprompt}{
    basicstyle=\ttfamily\scriptsize,
    columns=fullflexible,
    keepspaces=true,
    breaklines=true,
    breakatwhitespace=false,
    showstringspaces=false,
    frame=single,
    framerule=0.2pt,
    xleftmargin=0.4em,
    xrightmargin=0.2em,
    aboveskip=2pt,
    belowskip=2pt
}
\def\BibTeX{{\rm B\kern-.05em{\sc i\kern-.025em b}\kern-.08em
    T\kern-.1667em\lower.7ex\hbox{E}\kern-.125emX}}

\begin{document}

\title{Generate Then Correct: Single-Shot Global Correction for Aspect Sentiment Quad Prediction}

\author{Shidong He$^{1}$, Haoyu Wang$^{2}$, and Wenjie Luo$^{1*}$%
\thanks{*Corresponding author.}%
\thanks{$^{1}${\sloppy Shidong He and Wenjie Luo are with the School of Cyber Security and Computer, Hebei University, Baoding, China.
{\tt\small heshi\allowbreak dong@stumail.hbu.edu.cn, luowenjie@hbu.edu.cn}}}%
\thanks{$^{2}${\sloppy Haoyu Wang is with the Faculty of Computer Science, Dalhousie University, Halifax, Canada. {\tt\small hy873711@dal.ca}}}%
}

\maketitle

\begin{abstract}
Aspect-based sentiment analysis (ABSA) extracts aspect-level sentiment signals from user-generated text, supports product analytics, experience monitoring, and public-opinion tracking, and is central to fine-grained opinion mining. A key challenge in ABSA is aspect sentiment quad prediction (ASQP), which requires identifying four elements: the aspect term, the aspect category, the opinion term, and the sentiment polarity.
Existing generative methods linearize quads into a template and decode them in a single left-to-right pass, where early errors, once emitted, cannot be revisited.
By contrast, humans seldom expect to get structured outputs right on the first attempt, but instead draft first and then revise against the full context.
Following this intuition, we propose Generate-then-Correct (G2C): a generator drafts quads and a corrector performs a single-shot, sequence-level revision conditioned on the sentence and the draft, giving the model an explicit opportunity to fix near-miss errors from the first pass. The corrector is trained on LLM-synthesized drafts covering common error patterns.
On the Rest15 and Rest16 datasets, G2C outperforms strong baseline models. The code is available at https://github.com/EarthOnlinePlayer5732/G2C.
\end{abstract}

\section{Introduction}
Understanding not just whether a review is positive or negative, but exactly which aspects (e.g., \emph{food}, \emph{service}, \emph{price}) are praised or criticized, is essential for product analytics, customer-feedback monitoring, and public-opinion tracking.
Aspect-Based Sentiment Analysis (ABSA) addresses this need by decomposing an opinion into fine-grained elements, typically including aspect terms, aspect categories, opinion expressions, and sentiment polarities~\cite{liu-etal-2015-Aspect,zhou2015-cate,wang-etal-2016-Senti}.
Research has progressively moved from predicting these elements in isolation toward joint formulations that model their mutual dependencies~\cite{peng2020knowing,wan2020target}.

\begin{figure}[!t]
    \centering
    \includegraphics[width=1.00\linewidth]{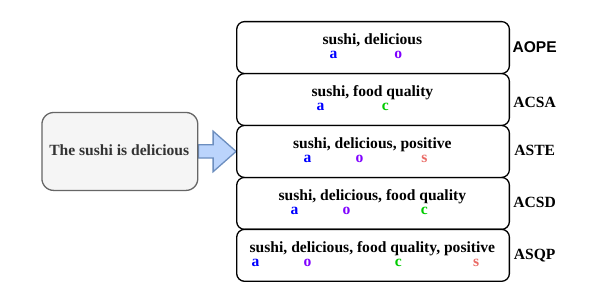}
    \caption{Differences Among ABSA Tasks.}
    \label{fig:TASK}
\end{figure}

Among joint formulations, aspect sentiment quad prediction (ASQP), also known as aspect-category--opinion--sentiment (ACOS)~\cite{cai-etal-2021-aspect}, has emerged as a particularly challenging task.
Given an input sentence, ASQP requires the model to output a set of quadruples $(a,c,o,s)$, where $a$ is the aspect term, $c$ is its aspect category from a predefined taxonomy, $o$ is the opinion term, and $s$ is the sentiment polarity.
This formulation unifies the core ABSA elements in a single structure and forces the model to decide how many quads are present, how aspects and opinions are paired, and which categories and polarities are appropriate for each pair.
Under the standard exact-match criterion, a predicted quad is counted as correct only when all four elements match the gold label; even a single-element deviation, however minor, results in a missed quad.

\begin{figure*}[t]
    \centering
    \includegraphics[width=1.00\textwidth]{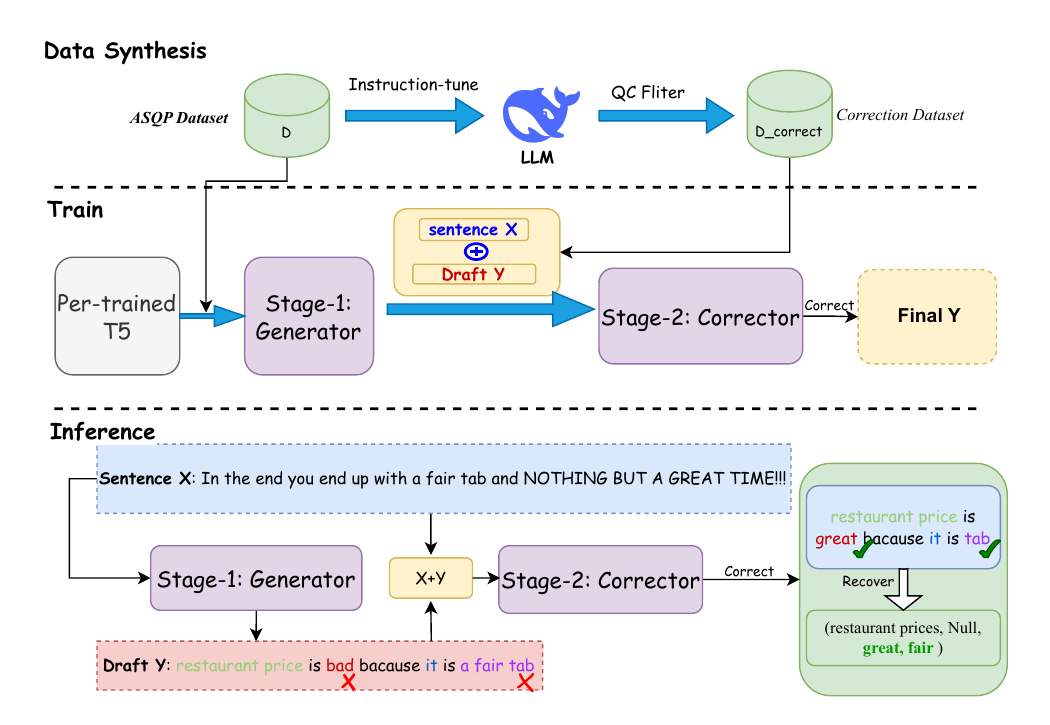}
    \caption{The proposed G2C framework.}
    \label{fig:framework}
\end{figure*}

Existing ASQP approaches mainly follow two paradigms.
Non-generative methods cast ASQP as tagging or classification, whereas generative methods decode a structured output with a pretrained sequence-to-sequence (seq2seq) model.
Because of their simplicity and end-to-end nature, generation-based methods have become dominant.
We follow this line and use the template \emph{[category] is [sentiment] because [aspect] is [opinion]}, where multiple quads are separated by \emph{[SSEP]}.
This linearization offers a unified text-to-text interface, but it requires the model to emit all elements in a single left-to-right pass. 
However, autoregressive decoding is prone to local errors such as a polarity flip, a shifted opinion span, or an aspect-opinion mispairing.
Because each token conditions every subsequent token, such errors can also affect later generation.
Crucially, single-pass decoding provides no mechanism to revisit these errors after they are emitted.

Prior work mitigates this from within the single-pass paradigm: order-selection and multi-view aggregation reduce sensitivity to a particular linearization order~\cite{hu-etal-2022-DLO/ILO,gou-etal-2023-mvp,jun-lee-2025-dynamic}, while modified training objectives suppress likely mistake tokens~\cite{hu-etal-2023-UAUL,zhu-etal-2024-GDP}.
These strategies improve the quality of the single pass itself, but the decoder still commits to each token as it is emitted; once generation finishes, there is no further stage that can revisit the output as a whole.

We take a different and complementary route.
In practice, humans seldom get a structured output right on the first attempt; they produce a complete draft and then revise it by re-reading the whole against the original input.
This draft-then-revise process is especially effective for near-miss errors---exactly the kind of local, single-element deviations that dominate generator output in ASQP.
Following this intuition, we propose Generate-then-Correct (G2C), a two-stage framework that keeps the standard text-to-text formulation without reverting to a pipeline~\cite{zhang-etal-2021-aspect-sentiment}.
Stage~1 performs single-pass decoding with the template to produce a complete draft.
Stage~2 uses a T5 model of the same architecture, initialized from the Stage-1 weights, to perform a one-shot, sequence-level revision conditioned on the sentence and the Stage-1 draft.
Because the corrector operates on a finished draft rather than a partial prefix, it can assess and fix errors with full-sequence context---an opportunity that no single-pass method provides.
To obtain training data for the corrector, we design an LLM-based synthesis pipeline that generates sentence--draft pairs with common near-miss error patterns, so that the corrector learns to map such noisy drafts to consistent and complete quad sets.

In summary, the main contributions of this paper are as follows:

(1) To the best of our knowledge, we are the first to introduce a two-stage Generate–then–Correct framework for ASQP, where a T5 corrector with the same architecture performs a \textbf{one-shot sequence-level revision} conditioned on the sentence and the Stage-1 draft, providing an explicit revision stage that single-pass decoding lacks.

(2) We develop an LLM-based synthesis pipeline that produces sentence--draft pairs with near-miss error patterns, providing dense supervision for training the corrector to map noisy drafts to consistent and complete quad sets.

(3) Experimental results show that on the Rest15 and Rest16 ASQP datasets, G2C achieves an average improvement of approximately 3.8\% over Stage-1 and also yields clear gains over previous strong baselines.

\section{Related Work}
\label{sec:related_work}

\textbf{Aspect-based sentiment analysis.}
ABSA has been widely studied for modeling fine-grained opinions toward aspects, evolving from predicting isolated elements to jointly extracting coupled opinion structures \cite{liu-etal-2015-Aspect,zhou2015-cate,wang-etal-2016-Senti,peng2020knowing,wan2020target}. In this paper, we focus on the ASQP task, i.e., quad-level joint extraction.

\textbf{Aspect sentiment quad prediction.}
ASQP extends ASTE by additionally predicting the aspect category, thereby requiring models to jointly model four coupled elements in each quad. Early non-generative approaches often decompose quad prediction into subtasks and solve them in a pipeline manner, which is conceptually simple but prone to error propagation. A representative instance is the extract-then-classify design that first identifies aspect–opinion pairs and then predicts their category and polarity \cite{cai-etal-2021-aspect}. Beyond pipelines, structured prediction/decoding has been explored to better preserve compositional relations, such as path-based structured generation \cite{mao-etal-2022-seq2path} and opinion-tree generation \cite{ijcai-OTG}. In parallel, generative approaches cast ASQP as text-to-text generation by linearizing quads into a template sequence and decoding it with a pretrained seq2seq model \cite{zhang-etal-2021-aspect-sentiment,zhang-etal-2021-GAS}.

Within the generative paradigm, robustness remains a recurring issue: autoregressive decoding is prone to local errors, and template design introduces order sensitivity~\cite{zhang-etal-2021-aspect-sentiment}.
Prior work mitigates these issues by decoding under multiple linearization orders and aggregating the results~\cite{hu-etal-2022-DLO/ILO,gou-etal-2023-mvp}, or by designing uncertainty-aware objectives to improve stability~\cite{hu-etal-2023-UAUL}.
Other methods explicitly target generation instability in quad decoding to enhance consistency of generated structures~\cite{zhu-etal-2024-GDP}.
Despite these advances, most methods still rely on a single left-to-right pass with limited opportunity to revise early decisions once emitted.

Our work is complementary: rather than improving the single pass itself, G2C keeps the standard generative formulation as Stage~1 and adds one corrector pass that revises the finished draft with full-sequence context.
The corrector is trained on LLM-synthesized near-miss drafts, which provide direct supervision for the correction mapping.

\section{Task Definition}
\label{sec:task}

Following prior work on generative ASQP~\cite{cai-etal-2021-aspect,zhang-etal-2021-aspect-sentiment}, we formulate aspect sentiment quad prediction as follows. Given an input sentence $x$, the goal is to predict a set of aspect-level quadruples
$\mathcal{Q}=\{(a_i,c_i,o_i,s_i)\}_{i=1}^{M}$.
Here $a_i$ denotes the aspect term (which may be implicit), $c_i$ is its aspect category chosen from a predefined taxonomy, $o_i$ is the opinion term, and $s_i\in\{\textit{positive},\textit{negative},\textit{neutral}\}$ is the sentiment polarity. The number of quads $M$ is not given in advance. The model must therefore determine how many quads are present in $x$, identify the spans of aspect and opinion terms, assign appropriate categories and polarities, and output the resulting quad set in a suitable textual form.

\section{Methodology}
\label{sec:method}

\subsection{Overview}
\label{ssec:overview}

To provide an explicit revision stage that single-pass decoding lacks, we propose a two-stage Generate-then-Correct framework.
As illustrated in Figure \ref{fig:framework}, G2C has two stages: (1) Stage‑1 (Generator), a T5 seq2seq model that drafts a set of quads under the linearization template; and (2) Stage‑2 (Corrector), a same‑architecture T5 initialized from Stage‑1 that performs a one‑shot sequence‑level rewrite conditioned on the sentence and the draft to produce the final quads. The Corrector is trained on LLM‑synthesized noisy drafts to learn a draft‑to‑gold mapping. At inference, we run one forward pass per stage with matched model sizes and decoding settings. This preserves the text‑to‑text formulation without reverting to a pipeline.

\subsection{Stage-1: Initial Prediction Generation}
\label{ssec:Generation}

The Stage-1 Generator is a T5-based\cite{T5} model fine-tuned to produce sentiment quads in a single pass. Given an input sentence, the model generates a textual draft encoding all predicted quadruples.
In the first stage, we employ a standard pre-trained T5 model as our Generator. Following the paradigm established by prior work\cite{zhang-etal-2021-aspect-sentiment}, the generator is fine-tuned on the standard ASQP training set to convert an input sentence $x$ into a natural language sequence $\tilde{y}$ that represents the predicted quadruples. This sequence $\tilde{y}$ serves as the initial, potentially flawed prediction.

We linearize a sentiment quad as follows.
\begin{equation}
q=(a,c,o,s).
\end{equation}
We verbalize it using the template below.
\begin{equation}
\text{[}c\text{] is [}s\text{] because [}a\text{] is [}o\text{]}.
\end{equation}
Here, \(c\) denotes the \textbf{category}, \(s\) the \textbf{sentiment}, \(a\) the \textbf{aspect term}, and \(o\) the \textbf{opinion term}. Multiple quads are separated by  \emph{[SSEP]}.

Let $x$ be a sentence and $\mathcal{Q}=\{(a_i,c_i,o_i,s_i)\}_{i=1}^{M}$ its quad set.
Let $\ell(\cdot)$ verbalize one quad using the above pattern.
We define the linearization as follows:
\begin{equation}
\label{eq:seq_build}
\scalebox{1.00}{$y = L(\mathcal{Q})
= \texttt{[SSEP]}\,\text{.join}\!\big(\{\ell(a_i,c_i,o_i,s_i)\}_{i=1}^{M}\big)
$}
\end{equation}

To strengthen task conditioning, we prepend a lightweight instruction prefix that specifies the quad schema and output formatting constraints.
The Generator is a T5 model trained with maximum-likelihood estimation to map an input sentence $x$ into a linearized draft $\tilde{y}$ that represents the predicted quads (i.e., $\tilde{y}=L(\widehat{\mathcal{Q}}^{(1)})$).

Formally, Stage-1 is a standard autoregressive seq2seq model trained with masked token-level cross-entropy.
\begin{equation}
\label{eq:gen_loss}
\mathcal{L}_{\text{gen}}=-\log p_{\theta}(\tilde{y}\mid x).
\end{equation}

\subsection{Stage-2: Correction}
\label{ssec:Correction}

The Stage-2 Corrector is a sequence-level model that performs \textbf{one-shot sequence-level global correction} conditioned on both the sentence and the Stage-1 draft. We use a T5 of the same architecture and size as the Generator and initialize it from the Generator’s weights.
The input concatenates the raw sentence $x$ and the draft $\tilde{y}$ with special delimiters, and the model produces the final linearized quad sequence in a single decoding pass. Formally, we denote the corrector’s conditional distribution as
\begin{equation}
\label{eq:corr_dist}
q_{\phi}(y\mid x,\tilde{y}).
\end{equation}

We fine-tune the Corrector by minimizing a weighted, masked negative log-likelihood over sentence–draft–gold triples $(x_i,\tilde{y}_i,y_i)$. 
For each example $i$, let $m_{i,t}\in\{0,1\}$ indicate supervised target positions and $T_i=\sum_t m_{i,t}$.
The objective is
\begin{equation}
\label{eq:corr_weighted_loss}\scalebox{1.00}{$
\mathcal{L}_{\text{corr}}(\phi)
=
-\frac{1}{N}\sum_{i=1}^{N}\frac{w_i}{T_i}
\sum_{t} m_{i,t}\,
\log q_{\phi}\!\left(y_{i,t}\mid y_{i,<t},x_i,\tilde{y}_i\right)$}
\end{equation}
where $q_{\phi}$ denotes the Corrector, and $w_i\!\in\!\{w_{\text{cor}},w_{\text{err}}\}$ optionally reweights identity pairs ($\tilde{y}_i\!\equiv\! y_i$) and error drafts; unless stated, $w_{\text{cor}}\!=\!w_{\text{err}}\!=\!1$.

This preserves a simple text-to-text interface. The corrector conditions on the complete draft rather than a partial prefix and is not tied to the original template order, so it can add, remove, or modify elements to better match the sentence, while stabilizing already-correct parts. By conditioning on $(x,\tilde{y})$, the Corrector reduces common near-miss errors (e.g., polarity flips, mild opinion-span drift, and (aspect, opinion) mispairings) and enforces sentence-level consistency.

\begin{table*}[t]
    \caption{Performance comparison of various methods on ASQP datasets. Baselines are reported as in their papers and all metrics are reported in percentages. The best results are in bold, while the second best are underlined}
    \label{tab:performance_comparison}
    \centering
    \small 
    \begin{tabular*}{0.85\textwidth}{l@{\extracolsep{\fill}}cccccc}
    \toprule
    \multirow{2}{*}{Method} & \multicolumn{3}{c}{Rest15} & \multicolumn{3}{c}{Rest16} \\
    \cmidrule(lr){2-4} \cmidrule(lr){5-7}
    & Pre & Rec & F1 & Pre & Rec & F1 \\
    \midrule
    HGCN-BERT+BERT-TFM(2021) & 25.55 & 22.01 & 23.65 & 27.40 & 26.41 & 26.90 \\
    HGCN-BERT+BERT-Linear(2021) & 24.32 & 20.25 & 22.15 & 25.36 & 24.03 & 24.68 \\
    Extract-Classify (2021) & 35.64 & 37.25 & 46.42 & 38.40 & 50.93 & 43.88 \\
    \midrule
    GAS (2021) & 45.31 & 46.70 & 45.98 & 54.54 & 57.62 & 56.04 \\
    Paraphrase (2021) & 46.16 & 47.72 & 46.93 & 56.63 & 59.30 & 57.93 \\
    DLO (2022) & 47.08 & 49.33 & 48.18 & 57.92 & 61.80 & 59.79 \\
    ILO (2022) & 47.78 & 50.38 & 49.05 & 57.58 & 61.17 & 59.32 \\
    Special\_Symbols (2022) & 48.24 & 48.93 & 48.58 & 58.74 & 60.35 & 59.53 \\
    DLO+UAUL (2023) & 48.03 & 50.54 & 49.26 & 59.02 & 62.05 & 60.50 \\
    Special\_Symbols+UAUL (2023) & 49.12 & 50.39 & 49.75 & 59.24 & 61.75 & 60.47 \\
    MVP (2023) & - & - & 51.04 & - & - & 60.39 \\
    GDP (2024) & 49.20 & 50.31 & 49.75 & 61.16 & 62.08 & 61.61 \\
    STAR (2024) & 50.80 & \underline{51.95} & 51.37 & 60.54 & 62.90 & 61.70 \\
    SCRAP (2024) & \textbf{55.45} & 45.41 & 49.93 & \textbf{69.59} & 56.70 & \underline{62.48}\\
    DOT (2025) & - & - & \underline{51.91} & - & - & 61.24 \\
    \midrule
    G2C (ours) & \underline{51.26} & \textbf{53.38} & \textbf{52.40}& \underline{61.24} & \textbf{64.48} & \textbf{62.80}\\
    \midrule
    - w/o Corrector (Stage-1)  & 48.62 & 50.94 & 49.75 & 59.90 & 62.83 & 61.33\\
    - w/o identify\_pairs & 49.15 & 51.19 & 50.15 & 60.29 & \underline{63.45} & 61.83\\
    \bottomrule
    \end{tabular*}
\end{table*}

\subsection{Data}
\label{ssec:data}

\begin{figure}
    \centering
    \includegraphics[width=1.00\linewidth]{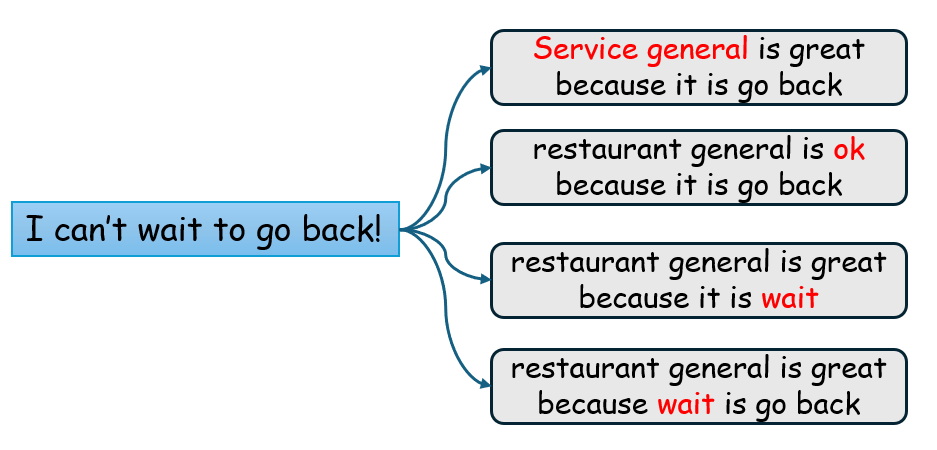}
    \caption{Example of an error.}
    \label{fig:error}
\end{figure}
Training the Corrector requires abundant paired examples of flawed drafts and gold outputs, which standard ASQP datasets do not provide. We therefore synthesize five draft variants per training sentence with an instruction‑tuned LLM (DeepSeek‑R1~\cite{deepseekr1}) in a single call;  a condensed illustration of the prompt is provided in the Appendix, and the full version is released with our code. Four drafts each perturb one quad component (Category, Aspect, Opinion, or Sentiment), and the fifth randomly selects one of these four types. Meanwhile, for each training sentence we also synthesize a corresponding identity pair ($\tilde{y}=y$).

As shown in Fig.~\ref{fig:error}, we generate drafts that intentionally introduce a single error in one component of the quad, covering four classes:
\begingroup
\renewcommand{\labelenumi}{(\arabic{enumi})}
\begin{enumerate}
\item \textbf{Aspect} --- replace the aspect term with a plausible but incorrect in-sentence mention, preferably semantically related and matching the noun or noun-phrase type;
\item \textbf{Category} --- substitute a semantically related but wrong label from the predefined taxonomy;
\item \textbf{Sentiment} --- change the sentiment verbalization to a different value;
\item \textbf{Opinion} --- replace the opinion term with a plausible but incorrect in-sentence mention, preserving part-of-speech or phrase type when possible.
\end{enumerate}
\endgroup
In sentences with multiple aspects or opinions, these perturbations can induce aspect-opinion mispairings or mild opinion-boundary drift.
We keep such cases because they mirror typical generator failures.

After synthesis, we apply quality‑control filtering: duplicates are removed; drafts that materially contradict the sentence are discarded; and only drafts that remain parseable under the linearization template are kept. Concretely, aspect/opinion replacements must come from the source sentence (surface form or substring), and category labels must belong to the predefined taxonomy. Each retained draft pairs with its sentence as input and $y$ as the target, yielding $(x,\tilde{y}) \to y$ instances.

\section{Experiments}
\label{sec:Experiments}

\subsection{Dataset}
\label{ssec:dataset}

We employ two restaurant-domain datasets for ASQP: Rest15 and Rest16, originally from the SemEval tasks~\cite{rest15,rest16} and later refined and extended by Zhang~\cite{zhang-etal-2021-aspect-sentiment}.
The split statistics of these two datasets are summarized in Table~\ref{tab:DATASET}.

\begin{table}[t]
    \caption{Statistics of Rest15 and Rest16 datasets.}
    \label{tab:DATASET}
    \centering
    \small
    \begin{tabular}{lccc}
    \toprule
    Dataset & Rest15 & Rest16 \\
    \midrule
    Train & 834 & 1264  \\
    Validation & 209 & 316 \\
    Test & 537 & 544  \\
    \bottomrule
    \end{tabular}
\end{table}
Model performance is reported precision (Pre, \%), recall (Rec, \%), and F1 (F1, \%) at the quad level. A sentiment quad prediction is counted as correct if and only if all the predicted elements are exactly the same as the gold labels.
To avoid test leakage, draft synthesis for Corrector training uses only the non‑test splits. All results are reported on the official test sets.

\subsection{Baseline and Training Settings}
\label{ssec:baseline}

\textbf{Baseline.} We compare against representative and strong generation-based baselines for ASQP, including:
\textbf{HGCN-BERT+BERT-Linear} and \textbf{HGCN-BERT+BERT-TFM} \cite{zhang-etal-2021-aspect-sentiment} are early pipeline baselines; 
\textbf{Extract-Classify}\cite{cai-etal-2021-aspect} adopts an extract-then-classify pipeline for ASQP;
\textbf{GAS}~\cite{zhang-etal-2021-GAS}, the first unified generative ABSA framework;
\textbf{PARAPHRASE}~\cite{zhang-etal-2021-aspect-sentiment}, solves ASQP via fixed-order paraphrase generation;
\textbf{DLO}~\cite{hu-etal-2022-DLO/ILO}, selects dataset-level template orders by minimal-entropy scoring for augmentation;
\textbf{ILO}~\cite{hu-etal-2022-DLO/ILO}, selects instance-level template orders by minimal-entropy scoring for augmentation;
\textbf{ILO+UAUL} and \textbf{DLO+UAUL}~\cite{hu-etal-2023-UAUL}, add uncertainty-aware unlikelihood learning to explicitly suppress mistake tokens;
\textbf{MVP}~\cite{gou-etal-2023-mvp}, prompts multiple element orders and selects tuples by voting;
\textbf{DOT}~\cite{jun-lee-2025-dynamic}, dynamically generates instance-specific order templates using only necessary views;
\textbf{GDP}~\cite{zhu-etal-2024-GDP}, combats generation instability using diffusion learning and consistency likelihood;
\textbf{STAR}~\cite{star}, a method for stepwise task augmentation and relation learning in ASQP;
and \textbf{SCRAP}~\cite{kim-etal-2024-SCRAP}, which generates reasoning plus quadruplets with consistency voting and an extract-then-assign strategy.

\textbf{Training settings.} We use T5-base\cite{T5} as the backbone and AdamW as the optimizer. Stage 1 is trained with a learning rate of 3e-4 for 20 epochs with a batch size of 16; Stage 2 uses a learning rate of 1e-4 for 25 epochs with a batch size of 16. All experiments are conducted on a single NVIDIA RTX 4090 GPU. The final model is the last-epoch checkpoint in each stage. Decoding settings are matched across Stage 1 and G2C.

\subsection{Result}
\label{ssec:result}
\textbf{Main comparisons.} Table~\ref{tab:performance_comparison} reports the results on Rest15 and Rest16. Our method outperforms the single-pass Generator (Stage-1) on both datasets.
On Rest15, F1 rises from 49.75 to 52.40 (+5.30\%), with precision 48.62 to 51.26 and recall 50.94 to 53.38. On Rest16, F1 improves from 61.33 to 62.80 (+2.39\%), with precision 59.90 to 61.24 and recall 62.38 to 64.46. These gains indicate that the corrective stage improves both coverage and exactness of predicted quads. 
Moreover, G2C also attains the best performance among recently published baselines on both Rest15 and Rest16.

\begin{figure}[h]
    \centering
    \includegraphics[width=1.00\linewidth]{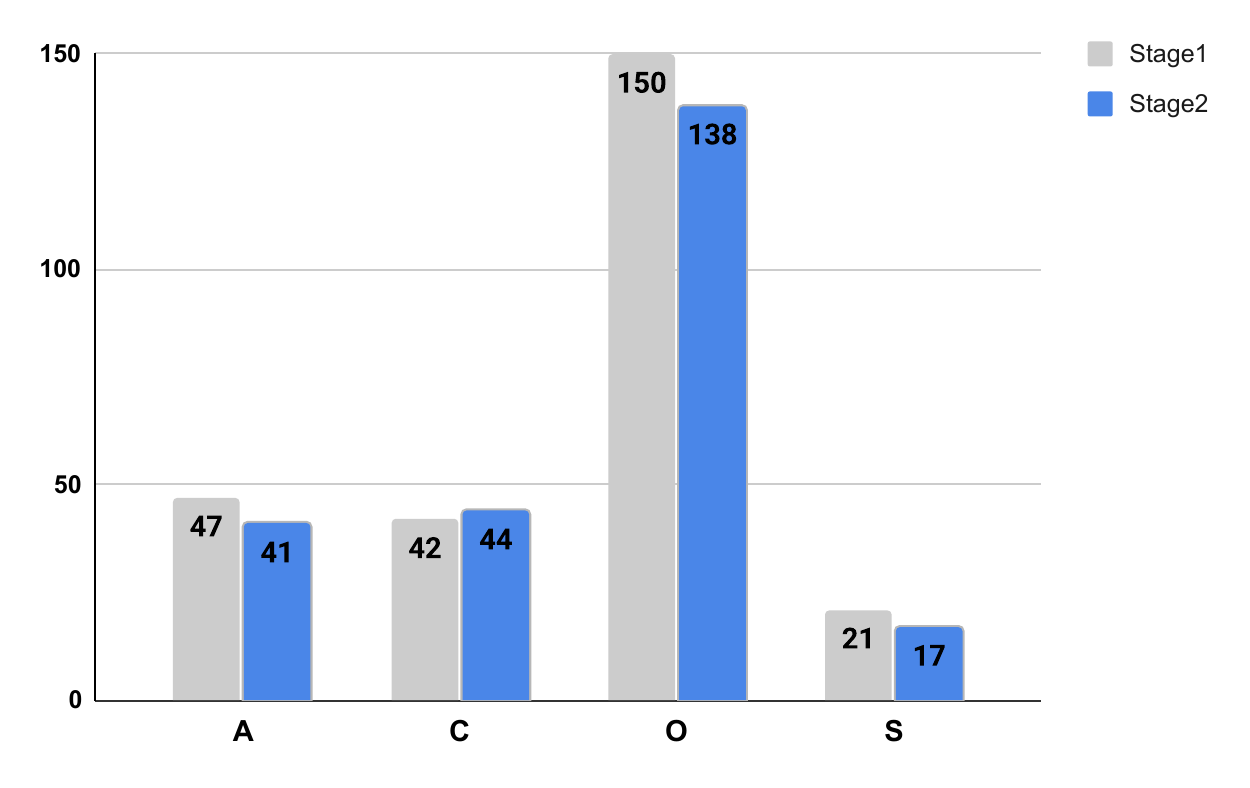}
    \caption{Single element error statistics in the Rest15 dataset.}
    \label{fig:analysis}
\end{figure}

\textbf{Analysis. }We further analyze G2C on Rest15. In Stage‑1, single‑element errors (predicted quads that differ from gold in exactly one element) account for more than half of all errors (260/430, 60.4\%). Figure~\ref{fig:analysis} shows that Stage-2 reduces several error types: opinion errors decrease from 150 to 138 (-8.0\%), sentiment errors from 21 to 17 (-19.0\%), and aspect errors from 47 to 41 (-12.7\%). This pattern indicates a consistent reduction across multiple single‑element error types rather than improvement in only one type. The slight rise in category single‑element errors (from 42 to 44) is due to error compression: some Stage‑1 multi‑element errors are partially fixed so that only the category remains wrong, migrating from multi‑element to category‑only errors. Overall, fewer wrong elements, fewer deletions, rare regressions, and the compression of complex errors into easier‑to‑localize category‑only errors are consistent with the concurrent gains in precision and recall. These results confirm that the revision stage is especially effective for the near-miss errors that motivate G2C.

\begin{table}[t]
    \caption{Performance comparison of inference time (seconds) on ASQP datasets.}
    \label{tab:inference_time}
    \centering
    \small
    \begin{tabular}{lccc}
    \toprule
    Method & Rest15 & Rest16 & Average \\
    \midrule
    Paraphrase (2021) & 10 & 13 & 11.5 \\
    MVP (2023) & 1932 & 1970 & 1932 \\
    DOT (2025) & 257 & 271 & 262 \\
    \midrule
    Ours & 70 & 74 & 72 \\
    \bottomrule
    \end{tabular}
\end{table}
\textbf{Ablation Study. }Table \ref{tab:performance_comparison} reports Rest15 and Rest16 for three variants: Stage 1 (Generator-only), G2C (full), and G2C without identity pairs. Removing the Corrector (Stage 1) leads to a clear drop on both datasets, with an average decrease of 3.6\%. Removing identity pairs also reduces F1; keeping a modest proportion during Corrector training is therefore beneficial. The full G2C achieves the best results.

\textbf{Inference Time Analysis. } Table~\ref{tab:inference_time} reports inference time comparisons of multiple methods on the ASQP benchmarks Rest15 and Rest16.
As shown in the table, although our method introduces an additional inference pass, it remains markedly more efficient than prior strong baselines, including the multi-view aggregation MVP and the DOT variant that adds dynamic ordering templates on top of MVP.
On Rest15 and Rest16, the inference times are 10 s and 13 s for Paraphrase, 1932 s and 1970 s for MVP, 257 s and 271 s for DOT, and 70 s and 74 s for Ours (G2C), respectively.
Note that multi-view methods such as MVP and DOT require many decoding passes to generate and aggregate candidates under different linearization orders, whereas G2C needs only two forward passes (one per stage) with matched model sizes.
Overall, relative to the prior strong baselines MVP and DOT, our method reduces inference time by about 96\% and 73\% on average, indicating a favorable efficiency–accuracy trade-off.

\section{CONCLUSION}
\label{sec:conclusion}

Single-pass generative ASQP provides no mechanism to revisit errors once emitted.
Borrowing from the human practice of drafting then revising, Generate-then-Correct (G2C) adds a one-shot sequence-level revision stage: a T5 corrector, trained on LLM-synthesized near-miss drafts, rewrites the Stage-1 output conditioned on the sentence and the draft while preserving the text-to-text interface.
On Rest15/Rest16, the corrector effectively fixes near-miss errors in the generator draft, and the resulting G2C framework surpasses recent strong baselines.
In future work, we will explore incorporating reasoning signals to further improve correction and consistency checking in challenging settings.

\appendix
\section{Synthesis Prompt}
\label{sec:appendix_prompt}

Below is a condensed illustration of the instruction prompt used to guide DeepSeek-R1 in synthesizing near-miss drafts. The full prompt is released with our code.
 
\begin{tcolorbox}[
  title={\small\bfseries LLM Synthesis Prompt},
  colback=white,
  colframe=black!75,
  coltitle=white,
  colbacktitle=black!75,
  breakable,
  fontupper=\footnotesize,
  left=5pt, right=5pt, top=4pt, bottom=4pt,
  boxrule=0.5pt
]
 
\textbf{Objective.}\enspace
Generate synthetic error correction data for Aspect Sentiment Quad Prediction.
 
\smallskip\noindent
\textbf{Core Task.}\enspace
For each input (a sentence + its correct quadruples), generate \textbf{6 lines}: 5 distinct error lines + 1 correct baseline, using the template and vocabulary below.
 
\smallskip\noindent\rule{\linewidth}{0.3pt}
\smallskip\noindent
\textbf{Output Format}
 
\smallskip\noindent
\textit{Lines 1--5 (Error Data):}\\
\texttt{Sentence [SENTSEP] <ModifiedQuads> \#\#\#\# <CorrectQuads>}
 
\noindent
\textit{Line 6 (Correct Baseline):}\\
\texttt{Sentence [SENTSEP] <CorrectQuads> \#\#\#\# <CorrectQuads>}
 
\noindent
Quadruples are separated by \texttt{[SSEP]}. In Lines 1--5, unmodified quadruples keep their correct form. Each of the 5 lines must introduce a \textbf{different} set of errors, varying the error type, the quadruple(s) modified, and the incorrect terms used.
 
\smallskip\noindent\rule{\linewidth}{0.3pt}
\smallskip\noindent
\textbf{Quadruple Template \& Vocabulary}
 
\smallskip\noindent
Template:\enspace\texttt{[category] is [sentiment] because [aspect] is [opinion]}\\
Sentiment:\enspace\texttt{great}\,(pos.),\enspace\texttt{bad}\,(neg.),\enspace\texttt{ok}\,(neu.)\\
Aspect ``NULL'' $\to$ render as ``it''.\\
Valid categories: location general, food prices, food quality, food general, ambience general, service general, restaurant prices, drinks prices, restaurant miscellaneous, drinks quality, drinks style\_options, restaurant general, food style\_options.
 
\smallskip\noindent\rule{\linewidth}{0.3pt}
\smallskip\noindent
\textbf{Error Types}\enspace(each modifies exactly one component)
 
\smallskip\noindent
\begin{tabular}{@{}l@{\enspace}p{0.82\linewidth}@{}}
\textit{Aspect} & Replace with a plausible but incorrect term from the sentence; prefer semantically related terms matching the grammatical role.\\[2pt]
\textit{Category} & Replace with a semantically related but incorrect category from the predefined list.\\[2pt]
\textit{Sentiment} & Replace with a different value from \{\texttt{great}, \texttt{bad}, \texttt{ok}\}; should represent a plausible misclassification.\\[2pt]
\textit{Opinion} & Replace with a plausible but incorrect term from the sentence; prefer semantically related terms.
\end{tabular}
 
\smallskip\noindent\rule{\linewidth}{0.3pt}
\smallskip\noindent
\textbf{Example}
 
\smallskip\noindent
\textit{Input:}\enspace Sentence = ``The sushi was fresh but overpriced.''\enspace Quads = [(food quality, positive, sushi, fresh), (food prices, negative, sushi, overpriced)]
 
\smallskip\noindent
\textit{Output (6 lines):}
 
\begin{lstlisting}[style=appendixprompt,numbers=left,numberstyle=\tiny,stepnumber=1,numbersep=4pt]
The sushi was fresh but overpriced. [SENTSEP] food general is great because sushi is fresh [SSEP] food prices is bad because sushi is overpriced #### food quality is great because sushi is fresh [SSEP] food prices is bad because sushi is overpriced
The sushi was fresh but overpriced. [SENTSEP] food quality is great because fresh is fresh [SSEP] food prices is bad because sushi is overpriced #### food quality is great because sushi is fresh [SSEP] food prices is bad because sushi is overpriced
The sushi was fresh but overpriced. [SENTSEP] food quality is ok because sushi is fresh [SSEP] food prices is bad because sushi is overpriced #### food quality is great because sushi is fresh [SSEP] food prices is bad because sushi is overpriced
The sushi was fresh but overpriced. [SENTSEP] food quality is great because sushi is fresh [SSEP] food prices is bad because sushi is fresh #### food quality is great because sushi is fresh [SSEP] food prices is bad because sushi is overpriced
The sushi was fresh but overpriced. [SENTSEP] food quality is great because sushi is fresh [SSEP] restaurant prices is bad because sushi is overpriced #### food quality is great because sushi is fresh [SSEP] food prices is bad because sushi is overpriced
The sushi was fresh but overpriced. [SENTSEP] food quality is great because sushi is fresh [SSEP] food prices is bad because sushi is overpriced #### food quality is great because sushi is fresh [SSEP] food prices is bad because sushi is overpriced
\end{lstlisting}
 
\smallskip\noindent
{\footnotesize Lines 1--5: category error, aspect error, sentiment error, opinion error, category error (2nd quad). Line 6: identity pair.}
 
\smallskip\noindent\rule{\linewidth}{0.3pt}
\smallskip\noindent
\textbf{Output Constraints.}\enspace
Output ONLY the 6 data lines. No text, headers, labels, numbering, or error descriptions. Use separators \texttt{[SENTSEP]}, \texttt{[SSEP]}, \texttt{\#\#\#\#} exactly as specified.
 
\end{tcolorbox}
 
\bibliographystyle{IEEEtran}
\bibliography{refs}
\end{document}